# MAFW: A Large-scale, Multi-modal, Compound Affective Database for Dynamic Facial Expression Recognition in the Wild

Yuanyuan Liu
School of Computer Science, China
University of Geosciences (Wuhan)
(CUG)
Wuhan, Hubei, China
liuyy@cug.edu.cn

Wei Dai
School of Computer Science, CUG
Wuhan, Hubei, China
daiw@cug.edu.cn

Chuanxu Feng
School of Computer Science, CUG
Wuhan, Hubei, China
fcxfcx@cug.edu.cn

Wenbin Wang
School of Computer Science, CUG
Wuhan, Hubei, China
wangwenbin@cug.edu.cn

Guanghao Yin
School of Computer Science, CUG
Wuhan, Hubei, China
ygh2@cug.edu.cn

Jiabei Zeng
Institute of Computing Technology,
Chinese Academy of Sciences
Beijing, China
jiabei.zeng@ict.ac.cn

Shiguang Shan[*]
Institute of Computing Technology,
Chinese Academy of Sciences
Beijing, China
sgshan@ict.ac.cn

## ABSTRACT

Dynamic facial expression recognition (FER) databases provide important data support for affective computing and applications. However, most FER databases are annotated with several basic mutually exclusive emotional categories and contain only one modality, *e.g.*, videos. The monotonous labels and modality cannot accurately imitate human emotions and fulfill applications in the real world. In this paper, we propose MAFW, a large-scale multi-modal compound affective database with 10,045 video-audio clips in the wild. Each clip is annotated with a compound emotional category and a couple of sentences that describe the subjects' affective behaviors in the clip. For the compound emotion annotation, each clip is categorized into one or more of the 11 widely-used emotions, *i.e.*, anger, disgust, fear, happiness, neutral, sadness, surprise, contempt, anxiety, helplessness, and disappointment. To ensure high quality of the labels, we filter out the unreliable annotations by an Expectation Maximization (EM) algorithm, and then obtain 11 single-label emotion categories and 32 multi-label emotion categories. To the best of our knowledge, MAFW is the first in-the-wild multi-modal database annotated with compound emotion annotations and emotion-related captions. Additionally, we also propose a novel Transformer-based expression snippet feature learning method to recognize the compound emotions leveraging the expression-change relations among different emotions and modalities. Extensive experiments on MAFW database show the advantages of the proposed method over other state-of-the-art methods for both uni- and multi-modal FER. Our MAFW database is publicly available from https://mafw-database.github.io/MAFW.

## 1 INTRODUCTION

In recent years, facial expression recognition (FER) has become a hot research topic in the fields of human-computer interaction (HCI) systems, multimedia analysis and processing, intelligent robots, and so on [4, 11, 14, 33]. Despite the progress, most the existing methods and databases are developed based on six basic emotions (*i.e.*, happiness, sadness, fear, surprise, disgust, and anger) proposed by P. Ekman [13] and contain only a single modality, *e.g.*, videos. Since the monotonous labels and modality are significantly different from the real-world human emotions in the wild, FER techniques are still far from the real-world applications [15, 24]. In order to enhance the real-world use of FER technology, it is essential to construct a sizable, in-the-wild dynamic affective database encompassing compound emotions and modalities.

[*]Corresponding author

Table 1: Summary of existing dynamic facial expression databases.

| Database | #Sample | Source | Expression annotation | Is in-the-wild? | #Annotation Times | Modality |
|---|---|---|---|---|---|---|
| CK+ [28] | 327 | Lab | 6 expressions+neutral and contempt | No | - | Video |
| MMI [31] | 2900 | Lab | 6 expressions+neutral | No | - | Video |
| BP4D [39] | 328 | Lab | 6 expressions+embarrassment and pain | No | - | Video&Audio |
| Aff-Wild2 [8] | 84 | Web & YouTube | 6 expressions+neutral | Yes | 3 | Video&Audio |
| AFEW 7.0 [7] | 1,809 | 54 movies | 6 expressions+neutral | Yes | 2 | Video&Audio |
| CAER [22] | 13,201 | 79 TV dramas | 6 expressions+neutral | Yes | 3 | Video&Audio |
| EmoVoxCeleb [1] | 22,496 | Interview videos from YouTube | 6 expressions+neutral and contempt | Yes | Auto | Video&Audio |
| DFEW [20] | 16,372 | 1500 movies | 6 expressions+neutral | Yes | 10 | Video&Audio |
| Our MAFW | 10,045 | 1,600 movies & TV dramas<br>20,000 short videos from reality shows, talk shows, news, etc<br>2,045 clips from [7], [20], and [1] | 11 single expressions<br>32 multiple expressions<br>emotional descriptive text | Yes | 11 | Video<br>Audio<br>Text |

Existing dynamic databases are classified into two categories based on the method of collection: laboratory-collected constrained databases and in-the-wild databases [24]. Table 1 reports existing dynamic FER databases and their information. Through event induction, the constrained databases, including CK+ [28], MMI [31], etc., record films of facial expression changes in the lab. These databases with single, limited, and consistent expression changes have seen substantial breakthroughs in FER technology, but they fall short in simulating the complex real-world human emotions. The in-the-wild databases, such as AFEW 7.0 [7] and DFEW [20], are constructed by crawling videos from movies and TV dramas. These databases closely reflect actual life, including a variety of contextual factors and spontaneous expressions. However, they still have the following limitations:

- The labels of the data are monotonous. As shown in Table 1, most existing databases are composed of seven or eight basic mutually exclusive emotional categories, *e.g.*, six basic expressions plus neutral or contempt. Many studies [10, 12, 32, 35, 40] have shown that people usually express multiple emotions simultaneously in real life, along with gestures and vocal changes.
- Video sources are relatively homogeneous and repetitive. As shown in Table 1, videos in CAER [22] and DFEW [20] are from 79 TV dramas and 1,500 movies, respectively, while EmoVoxCeleb [1] is collected from interview programs.
- The modality of the data is relatively monotonous. As shown in Table 1, most existing FER databases contain only video and audio modalities, and very few contain text modalities.

To overcome the above problems, we construct a large-scale compound affective database called MAFW with multiple modalities in the wild, which contains 10,045 video-audio clips. MAFW can be used as a new benchmark for researchers to develop and evaluate their methods for several FER tasks, such as multi-modal emotion recognition, cross-domain FER, emotion captioning, self-supervision FER, etc. Fig. 1 gives typical examples and the corresponding annotations in our MAFW database. Our MAFW has the following three advantages over the existing databases:

- Our MAFW is the first large-scale, multi-modal, multi-label affective database with 11 single expression categories, 32 multiple expression categories, and emotional descriptive texts. To obtain reliable and objective annotation, each clip in MAFW is independently labeled enough often as one or more of the 11 expression categories. Unreliable labels are then removed by an Expectation Maximization (EM) based reliability evaluation algorithm.
- Unlike most existing multi-modal FER databases that are only labeled with expression category tags, we also provide bilingual descriptive texts on facial expressions and emotions for videos in English and Chinese. The descriptive texts include the information on the environment, body movements, facial unit action, and other emotional elements that can be used for both video emotion captioning and FER.
- Compared to existing databases whose sources are mostly movies and TV shows, MAFW also includes short videos of reality TV, talk shows, news, variety shows, etc.

In addition to MAFW, we also propose a novel Transformer-based expression snippet feature learning method (T-ESFL) to effectively model subtle intra-snippet and inter-snippet expression movements for discovering movement-sensitive emotion representation, thus obtaining robust uni- and multi-modal FER. Furthermore, we establish four benchmark evaluation protocols for MAFW. Extensive experiments show the advantages of T-ESFL over other state-of-the-art deep learning methods, for both uni- and multi-modal FER.

## 2 RELATED WORK

**Constrained dynamic FER databases** The constrained databases are usually captured from a small group of individuals in a fixed indoor setting, with emotion frequently occurring during video viewing and event elicitation. For example, CK+ [28] collected six basic expressions from 123 individuals under laboratory conditions. BP4D [39] collected eight expressions from 41 individuals in eight different scenarios, including one-to-one interviews (evoking pleasure), suddenly hearing a voice (evoking surprise), and so on. Despite being spontaneous, the constrained expression databases are limited by a single environment, simple settings, the number of individuals, and the cost of production, making it difficult to simulate the real-world human emotions.

**In-the-wild dynamic FER databases** Dynamic FER databases in the wild are usually collected from online sources like TV episodes, movies, and other media. AFEW 7.0 [7] and DFEW [20] collect 1,800 and 16,372 facial expression video clips from movies, respectively.

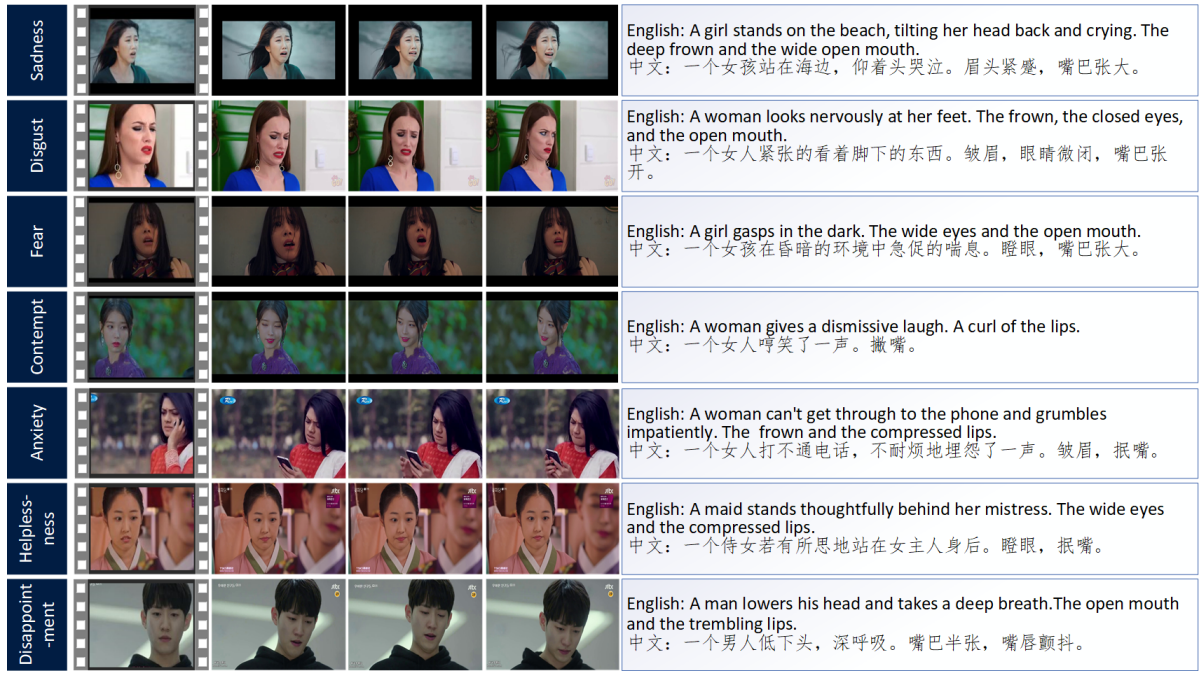

(a) Examples of the single expressions in MAFW.

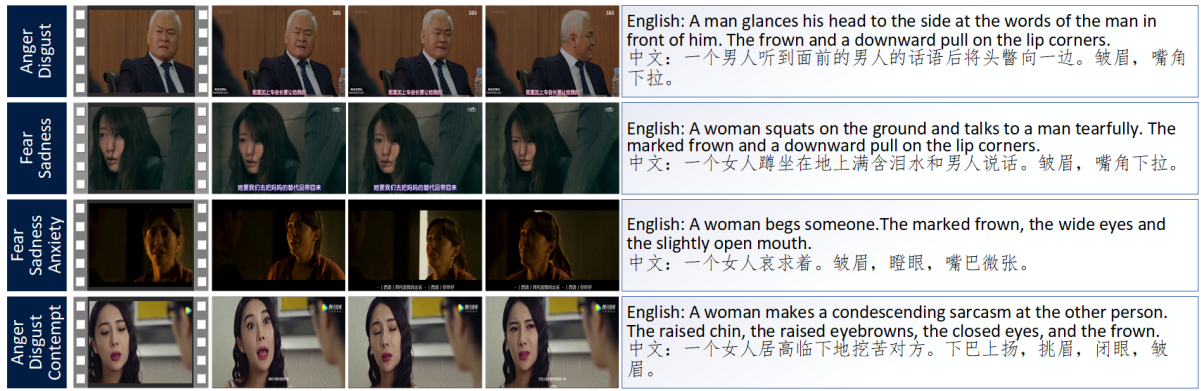

(b) Examples of the multiple expressions in MAFW.

Figure 1: Examples of the compound expressions and the bilingual descriptive texts from MAFW. (a) The single expressions in MAFW, (b) the multiple expressions in MAFW. Due to space limitations, we only show a small number of frames in these clips.

13,201 facial expression video clips from TV dramas are included in CAER [22]. Although these databases are created using real-world films, they all have the same restrictions, such as just offering basic and single expression labels and using movie or television clips as their sources.

**Compound FER databases** Recent studies in psychology and cognition have revealed that people frequently express compound emotions at once [12, 35]. This suggests that the existing FER databases with single, basic expression labels are not conducive to understand human emotions. In CVPR2017, Deng et al. [25] presented the first static compound FER database, namely RAF-DB, that contains 7-class single expressions and 12-class multiple expressions. In ACL2018, Zadeh et al. [2] presented a dynamic database, CMU-MOSEI, supporting multiple labels consisting of six basic expressions. Compared to these compound FER databases, our MAFW has more basic emotion categories, reliable multi-label emotion categories, and richer modalities.

## 3 MAFW DATABASE

### 3.1 Data Collection

The pipeline of data collection in MAFW is shown in Fig. 2. The MAFW has two main data sources. The first data sources are movies, TV dramas, and short videos from some reality shows, talk shows, news, variety shows, etc., on BiliBili and Youtube websites. We develop a crawler program to crawl over 1,600 HD movies, TV dramas, and over 20,000 short videos. These videos come from China, Japan, Korea, Europe, America, and India and cover various themes, *e.g.*, variety, family, science fiction, suspense, love, comedy, and interviews, encompassing a wide range of human emotions. To ensure the diversity of the data, we only randomly download one episode of the same TV series, as well as select no more than three facial expression clips in an episode or short video. The second data source, inspired by [38], uses videos from already-existing public databases to supplement some unusual categories, including 1,097 videos from DFEW [20], 98 videos from AFEW 7.0 [7], and 850 videos from EmoVoxCeleb [1].

With the crawled audio-video clips, we first use FaceDetector [18, 23] to detect the clips containing faces, then manually remove the unqualified clips to obtain 10,045 usable clips.

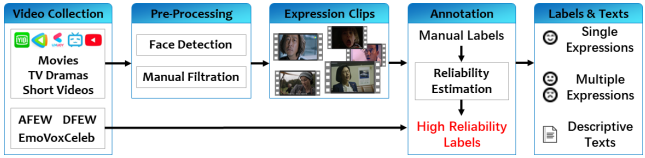

Figure 2: Overview of the construction of MAFW.

### 3.2 Data Annotation

Unlike other databases that only give the basic and single emotion annotation, our database offers three different types of emotion annotations for video clips: (1) **single expression label**, *i.e.*, each clip is assigned to a predominant and exclusive expression label, namely anger(AN), disgust(DI), fear(FE), happiness(HA), sadness(SA), surprise(SU), contempt(CO), anxiety(AX), helplessness(HL), disappointment(DS), and neutral(NE) (see Fig. 1(a)); (2) **multiple expression label**, *i.e.*, a clip can be annotated as a multi-label multiple expression category when it is determined to contain multiple emotions (such as "Anger+Disgust" in Fig. 1(b)); (3) **emotional descriptive text**, *i.e.*, each clip is bilingually annotated with a couple of sentences describing the subjects' affective behaviors in the clip. The following details the annotator selection, compound emotion category annotation, and descriptive text annotation, respectively.

**Annotator selection** Our annotators are college students from different degrees, majors, countries, and genders. To help annotate the emotion category and emotional descriptive text of each video, each annotator is initially trained to recognize expressions using the expression training tool mett[1] proposed by Paul Ekman to gain knowledge of facial action units and emotions. Following the instruction, the experts evaluate each annotator. Finally, for the annotation, 11 skilled annotators are used, each of whom had a test accuracy of at least 90%.

[1]https://www.paulekman.com/micro-expressions-training-tools/

**Compound emotion category annotation** To facilitate effective annotation, we create the ExpreLabelTool labeling tool. Each clip is categorized into one or more of the 11 complex emotions using the tool and is labeled by the 11 annotators. On a scale of 0 to 1, the annotators evaluate the self-confidence scores of their annotations (including 11 levels in ExpreLabelTool). The more certain the annotation is, the higher the score. After that, each clip can be obtained as an 11-dimensional vector, where each dimension represents the score of the labeled emotion. We describe later how to select *single* and *multiple* expressions based on this vector.

**Descriptive text annotation** For each video, except for the neutral emotion, the annotators are required to watch the video carefully and write down the bilingual emotional description according to the pre-established rules. Fig. 1 shows examples of the descriptive texts for emotion captioning in MAFW. The atmosphere, body movements, facial action units, and other emotional details are included in the captions. To ensure the complementarity of the emotional descriptive text, the descriptive text cannot directly use terms with emotional labels, such as "she is angry".

### 3.3 Metadata

The MAFW is a multi-modal database with text, audio, and video modalities. Each clip data is provided with a single or multiple expression label, an average confidence score for each emotion annotation, and several descriptive sentences (texts) for emotion captioning. We additionally offer three automatic annotations: the frame-level 68 facial landmarks, face regions, and gender. The gender of each person is identified by a CNN model that has been pre-trained on CelebA[26], and the facial landmarks and regions are detected by [3]. After identification and counting, 58.1% of the MAFW database is male and 41.9% is female.

### 3.4 Annotation Reliability Estimation

Due to the subjectivity difference of annotators, annotation reliability may be highly variable and inconsistent. To get rid of the labels with lower reliability, motivated by [34] and [5], we employ an Expectation Maximization (EM) algorithm to assess each annotator's reliability to achieve high-reliability labels. The algorithm of EM for reliability estimation is shown in Algorithm 1.

Given the labels of $N$ videos annotated by $M$ annotators, we first binarize their labels into a zero-one matrix $H_{MN}^k$ on the emotion category $k$ as:

$$H_{MN}^k = \{h_{ij}^k\}, \qquad (1)$$

where $h_{ij}^k$ will be "1", if the $i$th annotator labels the $j$th video with emotion category $k$, otherwise it will be "0".

Our goal is to estimate each annotator' reliability by optimizing the likelihood of their labels. The reliability is formulated as two M-dimensional probability vectors: $\{\alpha_i^k\}$ and $\{\beta_i^k\}$,

$$\alpha_i^k = P(h_{ij}^k = 1 | v_j^k = 1), \beta_i^k = P(h_{ij}^k = 0 | v_j^k = 0), \qquad (2)$$

where $\alpha_i^k$ is the reliability probability that the $i$th annotator correctly labels the emotion category $k$ and $\beta_i^k$ is the reliability probability that the $i$th annotator does not label the emotion category $k$. Note that $\alpha_i^k$ and $\beta_i^k$ are independent of each other. $v_j^k = \{0, 1\}$

denotes whether the $j$th video has the label of the emotion category $k$. We initialize the $v_j^k$ via annotation majority voting.

With the above definitions, in the E-step of the EM, the reliability probabilities are used to estimate the posterior probability $\varphi_j^k$ that the $j$th video correctly is labeled with the emotion category $k$:

$$\varphi_j^k = \frac{p^k \mu_j^k}{p^k \mu_j^k + (1-p^k)\eta_j^k}, \qquad (3)$$

where $p^k$ is the expected probability of the emotion category $k$ and initialized by $\frac{1}{N}\sum_{j=1}^{N} v_j^k$. $\mu_j^k$ and $\eta_j^k$ are calculated as:

$$\mu_j^k = \prod_{i=1}^{M}(\alpha_i^k)^{h_{ij}^k}(1-\alpha_i^k)^{(1-h_{ij}^k)}, \qquad (4)$$

$$\eta_j^k = \prod_{i=1}^{M}(\beta_i^k)^{(1-h_{ij}^k)}(1-\beta_i^k)^{h_{ij}^k}. \qquad (5)$$

In the M-step of the EM, we first update $p^k$ as:

$$p^k = \frac{1}{N}\sum_{j=1}^{N} \varphi_j^k. \qquad (6)$$

Then, we update $\alpha_i^k$ and $\beta_i^k$ by Maximum Likelihood Estimation:

$$\alpha_i^k = \frac{\sum_{j=1}^{N} \varphi_j^k h_{ij}^k}{\sum_{j=1}^{N} \varphi_j^k}, \qquad (7)$$

$$\beta_i^k = \frac{\sum_{j=1}^{N}(1-\varphi_j^k)(1-h_{ij}^k)}{\sum_{j=1}^{N}(1-\varphi_j^k)}. \qquad (8)$$

Finally, we set $Q(p^k, \alpha^k, \beta^k)$ as the convergence objective in EM algorithm as:

$$Q(p^k, \alpha^k, \beta^k) = \sum_{j=1}^{N}[\varphi_j^k \ln p^k \mu_j^k + (1-\varphi_j^k)\ln(1-p^k)\eta_j^k]. \qquad (9)$$

We can further determine whether $Q(p^k, \alpha^k, \beta^k)$ converges:

$$\frac{|Q(p_{(t+1)}^k, \alpha_{(t+1)}^k, \beta_{(t+1)}^k) - Q(p_{(t)}^k, \alpha_{(t)}^k, \beta_{(t)}^k)|}{|Q(p_{(t)}^k, \alpha_{(t)}^k, \beta_{(t)}^k)|} < \varepsilon, \qquad (10)$$

where $t$ denotes the number of iterations and $\varepsilon$ is the convergence threshold that is set as 0.000001 empirically. If $Q(p^k, \alpha^k, \beta^k)$ converges, we can obtain the reliability of all annotators, otherwise return the E-step.

**Table 2: Cronbach's alpha scores in the MAFW database.**

| Emotions | Alpha | Emotions | Alpha | Emotions | Alpha |
|---|---|---|---|---|---|
| Anger | 0.955 | Neutral | 0.878 | Anxiety | 0.729 |
| Disgust | 0.824 | Sadness | 0.948 | Helplessness | 0.686 |
| Fear | 0.934 | Surprise | 0.920 | Disappointment | 0.498 |
| Happiness | 0.961 | Contempt | 0.731 | **Average** | **0.824** |

With the reliability estimation, for each emotion category, we retain five high-reliability labels at least. We use Cronbach's Alpha [6] scores to measure the consistency of the retained labels. The results in Table 2 show that the retained labels have high consistency and reliability, with an average score of 0.823 on the 11-class emotion categories.

---

**Algorithm 1:** Annotation reliability estimation algorithm

**Input:**
zero-one matrix $\{H_{MN}^k\}_{k=1}^{K}$ of the emotion category $k$;
$M$: the number of annotators;
$N$: the number of videos;
$K$: the number of emotion categories.

**Output:** the reliability matrices of $M$ annotators on each emotion category $\{\alpha_i^k\}_{i=1}^{M}, \{\beta_i^k\}_{i=1}^{M}$.

**Initialize:**
$\forall k = 1, \ldots, K$, initialize true labels $\{v_j^k\}_{j=1}^{N}$ with majority voting via $H_{MN}^k$. The initial value of $p^k$ is the expected probability of the emotion label $k$.

$p^k := \frac{1}{N}\sum_{j=1}^{N} v_j^k \quad \alpha_i^k := 0.999999 \quad \beta_i^k := 0.999999$

**for** $k$=1 **to** $K$ **do**
  **Repeat**
    **E-step:**
      estimate the posterior probabilities $\{\varphi_j^k\}_{j=1}^{N}$ of $N$ clips with the $k$th expression as Eq. (3)–(5).
    **M-step:**
      update $p^k$, $\alpha_i^k$, and $\beta_i^k$ based on $\{\varphi_j^k\}_{j=1}^{N}$ through the maximum likelihood algorithm as Eq. (6)–(8).
      Calculate $Q(p^k, \alpha^k, \beta^k)$ as Eq. (9).
  **until** $Q(p^k, \alpha^k, \beta^k)$ **converges**

---

### 3.5 Single and Multiple Expression Selection

Using the retained high-reliability labels with their self-confidence scores, we can naturally divide the MAFW into two sets, namely the single expression set and the multiple expression set. Fig. 1(a) and Fig. 1(b) show some typical examples from the 11-class single expression and 32-class multiple expression sets, respectively.

Given the self-confidence scores from a high-reliability labeled clip, if no less than half of the annotators have labeled the $k$th emotion category $C^k = (c_1^k, c_2^k, \ldots, c_m^k)$, we then calculate the mean value of the self-confidence scores $c_{mean}^k = \sum_{i=1}^{m} c_i^k/m$ on the emotion category, and pick out the emotion label $k$ w.r.t $c_{mean}^k \geq 0.5$ as the valid label.

**Single expression set** For valid-labeled clips with single expression labels, we directly classify them into the single expression set; for clips with multiple expression labels, we select the labels with the highest average confidence score as its predominant single expressions and also classify them into the single expression set, so that the single expression set consists of all 9,172 valid-labeled clips with 11-class emotions. Table 3 reports the distribution of clip amount and clip length per expression category on the single expression set.

**Multiple expression set** Similarly, we create the multiple expression set from the valid-labeled clips with multiple expression labels. To prevent having too few samples in a class, we keep only the multiple expression categories with more than 10 labeled samples, yielding 32-class multiple expressions. As a result, we obtain 4,058 clips with multiple expressions. Fig. 3 shows the distribution of multiple expression categories on the multiple expression set.

**Table 3: The distribution of clip amount and clip length per single expression on the single expression set.**

| Expressions | Clips | | | | Percent(%) |
|---|---|---|---|---|---|
| | 0-2s | 2-5s | 5s+ | Total | |
| Anger | 183 | 945 | 262 | 1390 | 15.15 |
| Disgust | 97 | 434 | 108 | 639 | 6.97 |
| Fear | 139 | 413 | 73 | 625 | 6.81 |
| Happiness | 88 | 900 | 254 | 1242 | 13.54 |
| Neutral | 42 | 872 | 224 | 1138 | 12.41 |
| Sadness | 97 | 873 | 500 | 1470 | 16.03 |
| Surprise | 233 | 721 | 118 | 1072 | 11.69 |
| Contempt | 18 | 173 | 45 | 236 | 2.57 |
| Anxiety | 99 | 626 | 191 | 916 | 9.99 |
| Helplessness | 20 | 174 | 68 | 262 | 2.86 |
| Disappointment | 13 | 118 | 51 | 182 | 1.98 |
| Total | 1029 | 6249 | 1894 | 9172 | 100.00 |

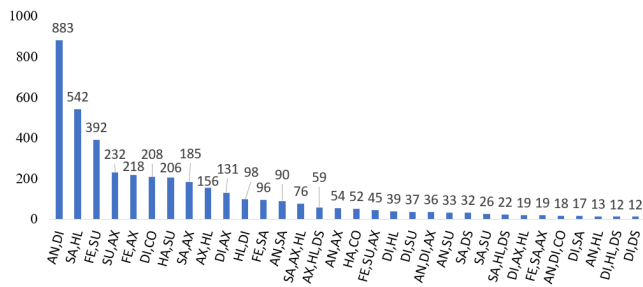

**Figure 3: The distribution of the number of multiple expressions on the multiple expression set.**

## 4 EXPRESSION SNIPPET FEATURE LEARNING WITH TRANSFORMER

In-the-wild FER is a difficult task due to subtle facial expression movements within videos that can be too difficult to be modeled properly by existing methods. In this paper, we propose a novel Transformer-based expression snippet feature learning method (T-ESFL) that can model intra-snippet and inter-snippet expression movements and relations, to obtain movement-sensitive emotion representation. In particular, for intra-snippet modeling, we decompose the modeling of facial movements of the entire video into the modeling of a series of small expression snippets so that enhance the encoding of subtle facial movements of each snippet by gradually attending to more salient information. Meanwhile, for inter-snippet modeling, we introduce a snippet order shuffling and reconstruction learning (SOSR) head and its loss to improve the modeling of subtle motion changes across snippets by training the Transformer to identify shuffled snippet orders. To this end, the

T-ESFL consists of three main components, *i.e.*, expression snippet decomposition, Transformer, and SOSR, as illustrated in Fig. 4.

**Expression snippet decomposition** Formally, given an input FER video clip $\mathcal{S}$, we first decompose the input into a series of small expression snippets $\mathcal{S} = \{S_1, S_2, ......S_n\}$, where $S_i$ represents the $i$-th snippet and $n$ is the total number of snippets. All the snippets have the same length, and they follow consecutive orders along time. To model subtle expression changes within each snippet, we employ a pre-trained CNN [29] and attention learning to extract snippet features $R_i$ from each $S_i$, thus augmenting the Transformer's ability to model intra-snippet expression changes.

**Transformer architecture** With the snippet features $R_i$, a Transformer is applied here to model the expression movements across snippets and discover a unified emotion feature for FER. We follow the typical Transformer [37] and apply a multi-head attention-based encoder-decoder pipeline for the processing. In general, the multi-head attention estimates the correlation between a *query* tensor and a *key* tensor and then aggregates a *value* tensor according to correlation results to obtain an attended output.

**SOSR learning** To make the output representation of the Transformer more sensitive to subtle expression movements, SOSR shuffles the snippet order and makes T-ESFL reconstruct the correct order in a self-supervision learning manner. The order of frames/audio within each snippet is retained. We follow a Jigsaw permutation [30] and shuffle the order pure randomly to deconstruct the normal temporal dependency between the snippets. The shuffled snippets are sent to T-ESFL and predicted the permutation type by using a reconstruction loss $L_{rec}$. Based on this, we can achieve movement-sensitive emotion representation $T$ for robust FER.

**Optimization Objective** The total objective function of T-ESFL includes two joint cross-entropy losses and is expressed as $L = L_{cls} + \frac{1}{n} \cdot L_{rec}$. The first one $L_{cls}$ is a FER classification loss, and the second one $L_{rec}$ is the snippet order reconstruction loss. Note that, $n$ is the number of the decomposed snippets.

**Multi-modal emotion prediction** The T-ESFL is easily extended for multi-modal FER, achieving the state-of-the-art performance on both uni- and multi-modal FER. Specifically, we use the ResNet_LSTM network and DPCNN [21] to extract audio and text emotion features, respectively. Then, we concatenate the audio, text, and movement-sensitive visual representations to identify the final emotion category via a simple fully-connected layer and Softmax operation. We experimentally verified that the use of multi-modal fusion features effectively improves FER in the wild.

## 5 EXPERIMENTS

The experimental setup of the benchmarks, including experiment protocols, data preprocessing, assessment measures, and implementation information, are first presented in the section. Then, using a variety of labels and modalities, we conducted comprehensive benchmarks and comparison studies on our MAFW.

### 5.1 Experimental Setup

**Data&Protocol** To facilitate the FER research from laboratory environments to the real world, we performed four challenging benchmark experiments on MAFW: 11-class uni-modal single expression classification, 11-class multi-modal single expression classification, 43-class uni-modal compound expression classification, and 43-class multi-modal compound expression classification. For 11-class single expression classification, the whole 9,172 clips from the entire single expression set were used to identify emotion categories. For the 43-class compound expression classification, which took into account both multiple and single expressions in real-world settings, 4,058 clips from the 32-class multiple expression set and the remaining 4,938 clips from the 11-class single expression set were used. Similar to the evaluation protocol of existing FER databases [20, 25], we adopt a 5-fold cross-validation protocol for these benchmarks on our MAFW database.

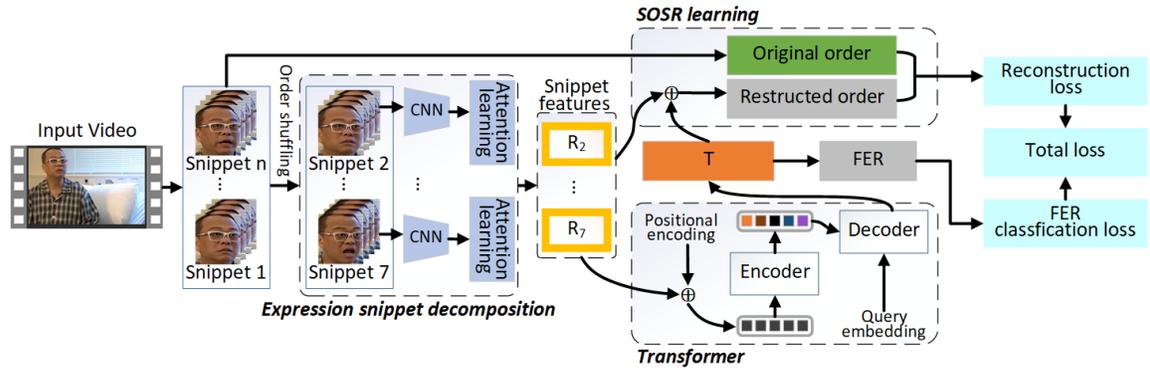

Figure 4: The architecture of T-ESFL for movement-sensitive emotion representation learning. Using untrimmed video clips, we mainly apply the expression snippet decomposition, the Transformer, and the SOSR, to enable the effective modeling of intra- and inter-snippet expression movements for discovering more informative expression cues, thus achieving robust FER.

**Preprocessing** First, we extracted frame pictures for each clip using OpenCV. Then, after deleting any frames without faces, we used the face-alignment-master program [3] to collect face areas and 68 landmarks on all frames. Finally, we performed face alignment using affine transform and matrix rotation in OpenCV.

**Evaluation Metrics** Consistent with the previous research [20, 25], we chose four widely-used validation metrics, *i.e.*, the unweighted average recall (UAR), weighted average recall (WAR), F-score (F1), and Area under the ROC curve (AUC), to evaluate the uni-modal and multi-modal FER tasks, respectively. The UAR is the average accuracy of all expression categories, regardless of the number of samples per class. The WAR is the recognition accuracy of overall expressions, which is related to the number of samples in each category. The F1 is regarded as the weighted harmonic mean value of the accuracy and recall, and here we simply calculate the average of the F1 on all categories. AUC generically refers to the area under the receiver operating characteristic (ROC) curve, and here we calculate the average AUC for all categories. We expect the proposed model to gain improvements in UAR, WAR, F1, and AUC metrics.

**Implementation Details** In this paper, we employed the PyTorch framework to implement all models. We conducted experiments in the uni-modal and multi-modal FER tasks, while each task contained single and compound expression classification, respectively. The key training parameters involved in the work are presented in Table 4. All models were trained on NVIDIA GeForce RTX 3090 and GTX1080, with an initial learning rate of 0.0001 provided by the grid search strategy. During training, the learning rate decreased at a rate of 0.2 when the loss was saturated.

Table 4: The key training parameters involved in the work.

| Models | Batch size | Input size |
| --- | --- | --- |
| Resnet18 [17], VIT [9] | 32 | 224 × 224 |
| C3D [36] | 8 | 112 × 112 |
| Resnet18_LSTM [16, 17, 19] | 16 | 224 × 224 |
| VIT_LSTM [9, 16, 19] | 16 | 224 × 224 |
| C3D_LSTM [16, 19, 36] | 8 | 112 × 112 |
| Resnet18_LSTM[a] [16, 17, 19] | 8 | 224 × 224 |
| C3D_LSTM[a] [16, 19, 36] | 8 | 112 × 112 |
| T-ESFL, T-ESFL[a], T-ESFL[a+t] | 8 | 224 × 224 |

[a] represents multi-modal evaluation with both video and audio;
[a+t] represents multi-modal evaluation with video, audio, and text.

### 5.2 Experimental Results

*5.2.1 11-class Uni-modal Single Expression Classification.* To evaluate uni-modal single expression classification, we compared our T-ESFL model with existing state-of-the-art FER models including three static frame-based methods ( *i.e.*, Resnet18 [17], VIT [9], and EmotionClassifier [18, 23]) and four dynamic sequence-based methods (*i.e.*, C3D [36], Resnet18 [16, 17, 19], VIT_LSTM [9, 16, 19], and C3D_LSTM [16, 19, 36]). The comparison results are shown in Table 5. For these static frame-based methods, we first selected five frames from a video evenly as input and then fused the prediction probabilities of the five frames in the output layer of the models to obtain the final prediction result. For these dynamic sequence-based methods, we used all frames in a video for emotion prediction. Compared to other state-of-the-art methods, the proposed T-ESFL achieved the best WAR of 48.18%. Moreover, our approach improved the WAR by 3.43% compared to the commercial model EmotionClassifier [18, 23], and also improved the WAR by 2.62% compared to the second best sequence-based method VIT_LSTM.

Table 5: Comparison results on 11-class uni-modal single expression classification.

| Models | Feature setting | AN | DI | FE | HA | NE | SA | SU | CO | AX | HL | DS | UAR | WAR |
|---|---|---|---|---|---|---|---|---|---|---|---|---|---|---|
| Resnet18 [17] | frame-based | 45.02 | 9.25 | 22.51 | 70.69 | 35.94 | 52.25 | 39.04 | 0 | 6.67 | 0 | 0 | 25.58 | 36.65 |
| VIT [9] | frame-based | 46.03 | **18.18** | 27.49 | 76.89 | 50.70 | 68.19 | 45.13 | 1.27 | 18.93 | 1.53 | 1.65 | 32.36 | 45.04 |
| EmotionClassifier [18, 23] | frame-based | 13.60 | 4.07 | 0.08 | 81.09 | **75.48** | 47.82 | 53.02 | - | - | - | - | **39.85** | 44.75 |
| C3D [36] | sequence-based | 51.47 | 10.66 | 24.66 | 70.64 | 43.81 | 55.04 | 46.61 | **1.68** | 24.34 | **5.73** | **4.93** | 31.17 | 42.25 |
| Resnet18_LSTM [16, 17, 19] | sequence-based | 46.25 | 4.70 | 25.56 | 68.92 | 44.99 | 51.91 | 45.88 | 1.69 | 15.75 | 1.53 | 1.65 | 28.08 | 39.38 |
| VIT_LSTM [9, 16, 19] | sequence-based | 42.42 | 14.58 | **35.69** | 76.25 | 54.48 | **68.87** | 41.01 | 0 | 24.40 | 0 | 1.65 | 32.67 | 45.56 |
| C3D_LSTM [16, 19, 36] | sequence-based | 54.91 | 0.47 | 9 | 73.43 | 41.39 | 64.92 | **58.43** | 0 | **24.62** | 0 | 0 | 29.75 | 43.76 |
| **T-ESFL** | snippet-based | **62.70** | 2.51 | 29.90 | **83.82** | 61.16 | 67.98 | 48.50 | 0 | 9.52 | 0 | 0 | 33.28 | **48.18** |

Table 6: Comparison results on 11-class multi-modal single expression classification.

| Models | Feature setting | AN | DI | FE | HA | NE | SA | SU | CO | AX | HL | DS | UAR | WAR |
|---|---|---|---|---|---|---|---|---|---|---|---|---|---|---|
| Resnet18_LSTM[a] [16, 17, 19, 27] | sequence-based | 54.47 | **11.89** | 7.07 | 82.73 | 54.85 | 55.06 | 39.35 | 0 | 15.99 | **0.39** | 0 | 29.26 | 42.69 |
| C3D_LSTM[a] [16, 19, 27, 36] | sequence-based | 62.47 | 3.17 | 15.74 | 77.30 | 42.20 | 65.30 | 42.67 | 0 | 19.14 | 0 | 0 | 30.47 | 44.15 |
| **T-ESFL[a]** | snippet-based | 60.73 | 1.26 | **21.4** | 80.31 | **58.24** | **75.31** | 53.23 | 0 | 14.93 | 0 | 0 | **33.35** | 48.7 |
| **T-ESFL[a+t]** | snippet-based | 61.89 | 1.1 | 7.69 | **85.90** | - | 71.87 | **62.17** | 0 | **36.00** | 0 | 0 | 31.00 | **50.29** |

[a] represents multi-modal evaluation with both video and audio;

[a+t] represents multi-modal evaluation with video, audio, and text.

### 5.2.2 11-class Multi-modal Single Expression Classification.

For multi-modal FER, we compared our T-ESFL model with two spatiotemporal neural network methods, *i.e.*, Resnet18_LSTM [16, 17, 19] and C3D_LSTM [16, 19, 36], as shown in Table 6. Obviously, the multiple modalities effectively improved the performance of FER. Compared to the other methods, our T-ESFL model obtained the best results in the fusion of different modalities, *e.g.*, 4.09% boost in UAR on video and audio modalities. Moreover, continuously adding the descriptive text modality obtained a relative 3.26% boost in WAR.

### 5.2.3 43-class Uni-modal Compound Expression Classification.

Table 7 shows the comparison results of 43-class uni-modal compound expression classification. Similar to the above single expression classification, the same six models except for the EmotionClassifier were used for 43-class uni-modal compound expression recognition, with the four evaluation metrics (WAR, UAR, F1, and AUC). Compared to the other methods, the proposed T-ESFL achieved the best WAR of 34.35% and the best AUC of 75.63%.

Table 7: Comparison results on 43-class uni-modal compound expression classification.

| Models | Feature setting | UAR | WAR | F1 | AUC |
|---|---|---|---|---|---|
| Resnet18 [17] | frame-based | 6.18 | 23.83 | 4.89 | 62.92 |
| VIT [9] | frame-based | 8.62 | 31.76 | 7.46 | 74.9 |
| C3D [36] | sequence-based | **9.51** | 28.12 | 6.73 | 74.54 |
| Resnet18_LSTM [16, 17, 19] | sequence-based | 6.93 | 26.6 | 5.56 | 68.86 |
| VIT_LSTM [9, 16, 19] | sequence-based | 8.72 | 32.24 | **7.59** | 75.33 |
| C3D_LSTM [16, 19, 36] | sequence-based | 7.34 | 28.19 | 5.67 | 65.65 |
| **T-ESFL** | snippet-based | 9.15 | **34.35** | 7.18 | **75.63** |

Table 8: Comparison results on 43-class multi-modal compound expression classification.

| Models | Feature setting | UAR | WAR | F1 | AUC |
|---|---|---|---|---|---|
| Resnet18_LSTM[a] [16, 17, 19, 27] | sequence-based | 7.85 | 31.03 | 5.95 | 71.08 |
| C3D_LSTM[a] [16, 19, 27, 36] | sequence-based | 7.45 | 29.88 | 5.76 | 68.13 |
| **T-ESFL[a]** | snippet-based | **9.93** | 34.67 | 8.44 | 74.13 |
| **T-ESFL[a+t]** | snippet-based | 9.68 | **35.02** | **8.65** | **74.35** |

[a] represents multi-modal evaluation with both video and audio;

[a+t] represents multi-modal evaluation with video, audio, and text.

### 5.2.4 43-class Multi-modal Compound Expression Classification.

For 43-class multi-modal compound expression classification, we also compared our T-ESFL with Resnet18_LSTM and C3D_LSTM, as shown in Table 8. Compared to the two methods on the multi-modal task, the proposed T-ESFL on video and audio modalities achieved the best UAR of 9.93% and WAR of 34.67%, respectively. Moreover, the results of T-ESFL kept achieving improvements after adding the descriptive text modality, *i.e.*, with a relative increase of 1% in WAR, 2.5% in F1, and 0.3% in AUC.

## 6 CONCLUSIONS AND FUTURE WORK

In this paper, we propose a large-scale, multi-label, multi-modal affective database called MAFW in the wild, which contains 10,045 video-audio clips. Each clip is annotated with a high-reliability compound emotional category and a couple of sentences that describe the subjects' affective behaviors in the clip. Therefore, MAFW is the first affective database that provides three types of emotion annotations, *i.e.*, single expression labels (11 class), multiple expression labels (32 class), and bilingual emotion captions. Moreover, we also propose a novel Transformer-based expression snippet feature learning method to obtain movement-sensitive emotion representation, thus achieving state-of-the-art performance on both uni-modal and multi-modal FER in the wild. In the future, we will continue to maintain the MAFW and hope that the release of this database can encourage more research on dynamic FER under unconstrained conditions, e.g., multi-modal emotion recognition, self-supervision FER, video emotion caption, zero-shot AU detection, etc.

## ACKNOWLEDGMENTS

This work is supported by the National Natural Science Foundation of China (Grant No. 62076227) and the Wuhan Applied Fundamental Frontier Project Grant (2020010601012166).

# A ADDITIONAL DETAILS IN MAFW

## A.1 Multiple Expression Distribution

Table 1 shows the distributions of 32-class multiple emotion categories on the multiple expression set, including the clip length and clip amount.

Table 1: The distributions of clip amount and clip length per multiple emotion category on the multiple expression subset.

| Expressions | Clips | | | | Percent(%) |
|---|---|---|---|---|---|
| | 0-2s | 2-5s | 5s+ | Total | |
| Anger,Disgust | 125 | 601 | 157 | 883 | 21.76 |
| Sadness,Helplessness | 21 | 307 | 214 | 542 | 13.36 |
| Fear,Surprise | 112 | 252 | 28 | 392 | 9.66 |
| Surprise,Anxiety | 36 | 173 | 23 | 232 | 5.72 |
| Fear,Anxiety | 37 | 146 | 35 | 218 | 5.37 |
| Disgust,Contempt | 26 | 151 | 31 | 208 | 5.13 |
| Happiness,Surprise | 23 | 152 | 31 | 206 | 5.08 |
| Sadness,Anxiety | 21 | 113 | 51 | 185 | 4.56 |
| Anxiety,Helplessness | 14 | 104 | 38 | 156 | 3.84 |
| Disgust,Anxiety | 16 | 96 | 19 | 131 | 3.23 |
| Helplessness,Disappointment | 10 | 64 | 24 | 98 | 2.41 |
| Fear,Sadness | 13 | 69 | 14 | 96 | 2.37 |
| Anger,Sadness | 9 | 55 | 26 | 90 | 2.22 |
| Sadness,Anxiety,Helplessness | 3 | 47 | 26 | 76 | 1.87 |
| Anxiety,Helplessness,Disappointment | 4 | 39 | 16 | 59 | 1.45 |
| Anger,Anxiety | 7 | 38 | 9 | 54 | 1.33 |
| Happiness,Contempt | 3 | 42 | 7 | 52 | 1.28 |
| Fear,Surprise,Anxiety | 8 | 27 | 10 | 45 | 1.11 |
| Disgust,Helplessness | 5 | 24 | 10 | 39 | 0.96 |
| Disgust,Surprise | 13 | 20 | 4 | 37 | 0.91 |
| Anger,Disgust,Anxiety | 3 | 31 | 2 | 36 | 0.89 |
| Anger,Surprise | 6 | 25 | 2 | 33 | 0.81 |
| Sadness,Disappointment | 2 | 21 | 9 | 32 | 0.79 |
| Sadness,Surprise | 1 | 14 | 11 | 26 | 0.64 |
| Sadness,Helplessness,Disappointment | 0 | 11 | 11 | 22 | 0.54 |
| Fear,Sadness,Anxiety | 4 | 13 | 2 | 19 | 0.47 |
| Disgust,Anxiety,Helplessness | 4 | 13 | 2 | 19 | 0.47 |
| Anger,Disgust,Contempt | 2 | 11 | 5 | 18 | 0.44 |
| Disgust,Sadness | 2 | 12 | 3 | 17 | 0.42 |
| Anger,Helplessness | 3 | 7 | 3 | 13 | 0.32 |
| Disgust,Disappointment | 3 | 7 | 2 | 12 | 0.30 |
| Disgust,Helplessness,Disappointment | 3 | 9 | 0 | 12 | 0.30 |
| Total | 539 | 2694 | 825 | 4058 | 100.00 |

## A.2 Annotation Format of the Compound Emotion

To efficiently annotate compound emotions, we developed an annotation tool called ExprLabelTool to generate and save annotation files for each annotator. Fig. 1 shows an annotated file format of a video-audio clip in MAFW. The "video_id" represents the index of the video-audio clip, the "labels" represents the expression categories labeled by an annotator for the clip, and the "scores" represents the self-confidence scores corresponding to the expression categories.

```
{
    "video_id" : 05237.mp4,
    "labels"   : [Disgust, Contempt],
    "scores"   : [0.6, 1.0]
}
```

Figure 1: An example of the emotion annotation file in MAFW.

## A.3 Annotation Format of the Emotional Descriptive Text

We carefully design our caption annotation task for emotional descriptive texts and develop several rules to ensure the sentences are of high syntactic and semantic quality in MAFW. Table. 2 shows the annotation instructions given to the annotators for the emotional description text.

Table 2: The annotation instructions given to the annotator for the emotional description text.

| Task |
|---|
| The task is to describe the emotional elements and the movements of the five facial features of the only main character in the video. The emotional elements include the body actions, the environment, the persons the character is speaking to, the tone of voice, and the the events' context. |
| **DOs** |
| 1. Each emotional description text is available in both Chinese and English. |
| 2. Use a personal pronoun as the subject of the sentence to refer to the main character in the video, such as "an old man", "a boy", etc., rather than their names (either the character's name or the actor's name). |
| 3. Use the simple present tense. |
| 4. Try to describe the part of the emotional elements in one sentence and modify the verb with an appropriate adverb to emphasize the sentiment state of the character. This part should be no less than eight words. |
| 5. Use predefined sentences to describe the part of the five facial features without adding new descriptions arbitrarily. |
| 6. Each sentence should be grammatically correct. |
| 7. It should describe only the visual content in the video. |
| **DONTs** |
| 1. Words that directly specify the expression category, such as "angry/anger/angrily", "sad/sadness/sadly", etc., should **NOT** appear. |
| 2. It should **NOT** describe your opinions, guesses or subjective judgements. |
| 3. It should **NOT** contain any digits. |

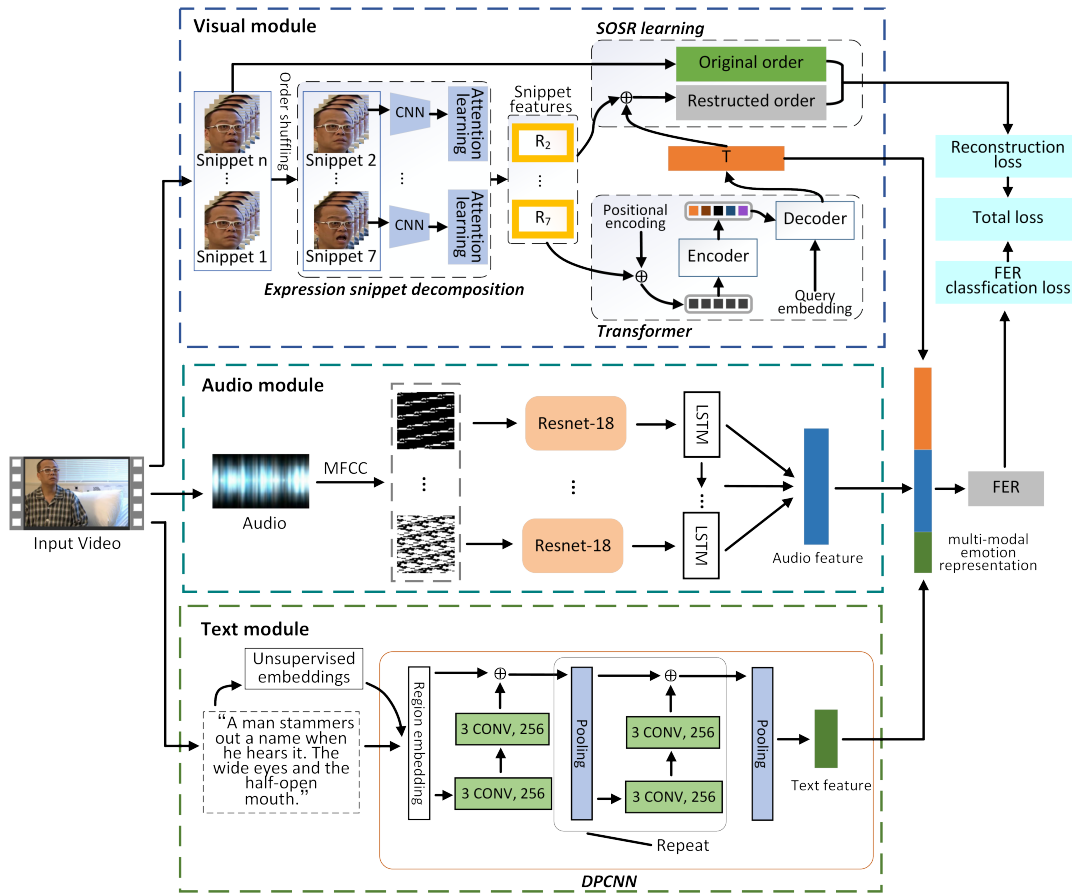

Figure 2: The framework with the T-ESFL for multi-modal emotion recognition.

## B THE DETAILED FRAMEWORK FOR MULTI-MODAL EMOTION RECOGNITION

The detailed framework with the T-ESFL for multi-modal emotion recognition is shown in Fig. 2. The multi-modal T-ESFL consists of three main modules, namely the visual module, the audio module, and the text module. First, the visual module uses snippet-based Transformer and SSOR to obtain the salient emotion feature $T$, the audio module uses ResNet_LSTM [2–4] to extract the audio emotion feature, and the text module uses DPCNN [5] to extract the text emotion feature. Then, we concatenate the visual, audio, and text features to generate the multi-modal emotion representation. As with the single-modal T-ESFL, the total objective function in the multi-modal emotion recognition includes cross-entropy loss and the snippet order reconstruction loss.

## C EXPERIMENTS FOR VIDEO EMOTIONAL CAPTIONING

We further discuss the application of our database to another task, e.g., video emotional captioning. To this end, we used two off-the-shelf video captioning models, namely Reconstruction network[10] and Video paragraph captioning model [8], to perform the video emotional captioning task and generate emotional text descriptions.

Table 3: Performance evaluation of video emotional captioning on our MAFW.

| Model | BLEU-4 | METEOR | CIDEr |
|---|---|---|---|
| Reconstruction network [10] | 6.87 | 11.50 | 25.63 |
| Video paragraph captioning model [8] | 9.09 | 15.49 | 23.40 |

We used three widely-used standard metrics in video captioning to evaluate the generated emotional text descriptions, namely BLEU-4[7], METEOR[1], and CIDEr[9]. Table 3 shows the experimental results of video emotional captioning using these two models in our database. Additionally, qualitative examples for video emotional captioning are shown in Figure 3.

## D ETHICAL STATEMENT

Although this is a purely academic investigation, the potential sensitivity of facial information necessitates an explicit statement of the ethics involved.

**Privacy**. Our method is used to capture features of facial expressions shared by many individuals, which are related to the common human perception of expressions. Therefore our method

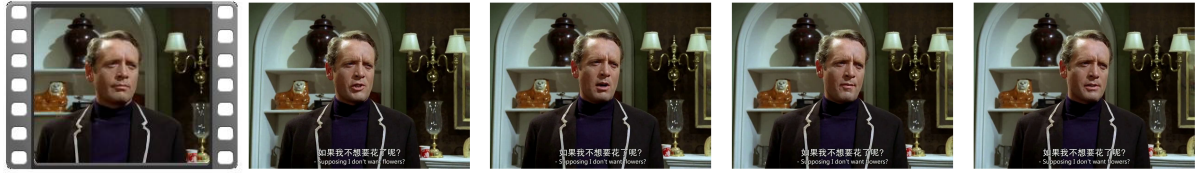

Reconstruction network: A man speaks to the camera, a man talks loudly
Video paragraph captioning model: A man speaks to the man in front of him The slight frown
Ground Truth: A man talks loudly. The tight frown and a downward pull on the right lip corner.

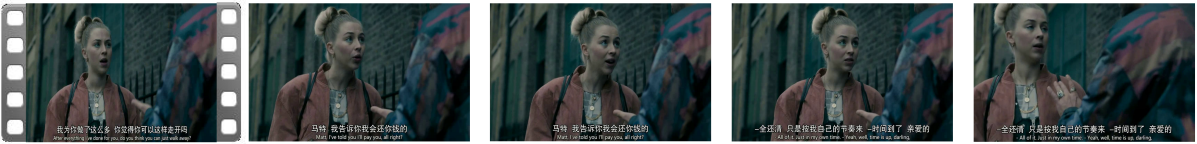

Reconstruction network: A woman speaks to the man in front of her, a woman talks to a man and expresses her displeasure
Video paragraph captioning model: A woman is not satisfied with the man in front of her The wide eyes the wrinkled nose
Ground Truth: A woman talks to a man and expresses her displeasure. The wide eyes, the higher inner corners of eyebrows and the lower outer corners of eyebrows.

**Figure 3: Visualization examples of video emotional captioning. The words in red are the predicted results of each model close to the Ground Truth, and the words in green are the Ground Truth.**

does not produce individual-specific facial expression analysis. Our MAFW database is used for academic research only and is compliant with GDPR [1] principles. The copyright of the original and cropped versions of the video remains with the original owner. No commercialization, secondary distribution or alteration of MAFW is allowed by any applicant.

**Database Bias**. During the data collection process, we did not differentiate any factors like gender, race, geography, age, etc. However, some data bias may occur in our MAFW database due to objective limitations such as data sources, the difficulty of collecting different emotions, etc.

**Metadata**. In our MAFW metadata, we use only the gender statistics automatically inferred from the model pre-trained on CelebA[6]. We only use this information to evaluate the distribution of data in our MAFW database and do not make use of it in our experiments or elsewhere.

[1]https://gdpr-info.eu/